\documentclass[runningheads]{llncs}
\usepackage[T1]{fontenc}
\usepackage{orcidlink}

\usepackage{amsmath}
\usepackage{amssymb}
\usepackage{amsfonts}

\usepackage{tabularx}

\usepackage{graphicx}    
\usepackage{booktabs}    
\usepackage{multirow}    
\usepackage{hyperref}    
\usepackage{url}         
\usepackage{algorithm}   
\usepackage{algorithmic} 
\usepackage{xcolor}      
\usepackage{fontawesome5}
\usepackage{mathtools}      
\usepackage{bm}             
\usepackage{bbm}            
\usepackage{enumitem}       
\usepackage{cleveref}       
\usepackage{annotate-equations}
\hypersetup{hidelinks}
\usepackage{color}

\graphicspath{{images/}}

\begin{document}
\title{\textbf{\sffamily{CoTu at EXACT 2026: Neuro-Symbolic Reasoning for Transparent Educational QA}}}
\titlerunning{\textbf{\sffamily{CoTu at EXACT 2026}}}
%

\newcommand{\bsf}[1]{\textbf{\sffamily{#1}}}
\newcommand*\samethanks[1][\value{footnote}]{\footnotemark[#1]}

\newcommand{\cotu}{\textbf{\sffamily{CoTu}}}
\newcommand{\backbone}{GRM-2.5}
\newcommand{\backboneHF}{\text{OrionLLM/GRM-2.5}}
\newcommand{\sftOne}{\text{CoTuGRM-2.5.T1-SFT}}
\newcommand{\sftTwo}{\text{CoTuGRM-2.5.T2-SFT}}

\author{Quoc-Khang Tran~\faUserTie~\orcidlink{0009-0000-2715-7647}\inst{1}\thanks{Corresponding author. Email: \{tqkhang,pnkhang,mtthanh\}@ctu.edu.vn} \and Minh-Thien Nguyen\orcidlink{0009-0004-1472-9537}\inst{1} \and Phu-An Thai\inst{1} \and Xuan-Tung Bui\orcidlink{0009-0005-2028-1035}\inst{1,2} \and Truong-Thanh Ma\inst{1}\samethanks \and Nguyen-Khang Pham\orcidlink{https://orcid.org/0009-0005-6761-1636}\inst{1}\samethanks}
%
\authorrunning{Quoc-Khang et al.}
%
\institute{Can Tho University, Can Tho city, Vietnam\\
\email{\{tqkhang,mtthanh,pnkhang\}@ctu.edu.vn,\\\{minhnguyent546,contact.thaiphuan\}@gmail.com,\\
tungp2425004@gstudent.ctu.edu.vn
}
\and Tay Do University, Can Tho city, Vietnam\\ \email{bxtung@tdu.edu.vn}}

\maketitle              
\begin{abstract}

Transparent educational question answering asks for answers that are not only correct but
explainable, and doing so with small models rules out the reasoning power of the largest proprietary
systems. The EXACT 2026 competition poses this problem concretely: open-weight language models of at
most 8B parameters, self-hosted, with a natural-language explanation for every answer. It pairs two
tasks: logical reasoning over university regulations, and multi-step physics problem solving. We describe
the system that team \cotu{} developed to address both, a neuro-symbolic
Program-of-Thought pipeline in which a 4B backbone writes a program rather than stating an answer
directly: for regulation queries it emits a Z3 encoding whose entailment verdict grounds the
deduction, and for physics it emits numerical Python, both wrapped in a shared self-correction loop
and a unified explained-JSON output. Answer-type routing, distillation-based task fine-tuning, and a
latency-aware serving stack---SGLang with speculative decoding---keep the system within the 60-second
per-query limit. The system achieved a \textbf{perfect score} on the physics task in both automated selection
rounds and obtained the \textbf{highest final-round technical score} of any team---$13.44/15$, combining
automated answer evaluation with expert-judged reasoning depth---with the equally weighted
presentation score included, \cotu{} placed 3rd overall. Grounding answers in a symbolic solver
yields correct, verifiable deductions at the 4B scale, and the residual difficulty lies in premise
selection rather than the deduction itself.

\vspace{0.1em}
\begin{center}
\parbox{0.03\textwidth}{\includegraphics[width=\linewidth]{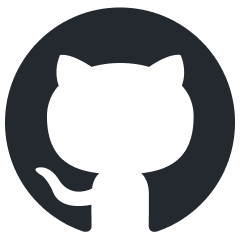}}\hspace{0.5mm}\href{https://github.com/minhnguyent546/EXACT-2026-CoTu}{\hspace{1mm}\texttt{EXACT-2026-CoTu}} \hspace{2pt}
\parbox{0.03\textwidth}{\includegraphics[width=\linewidth]{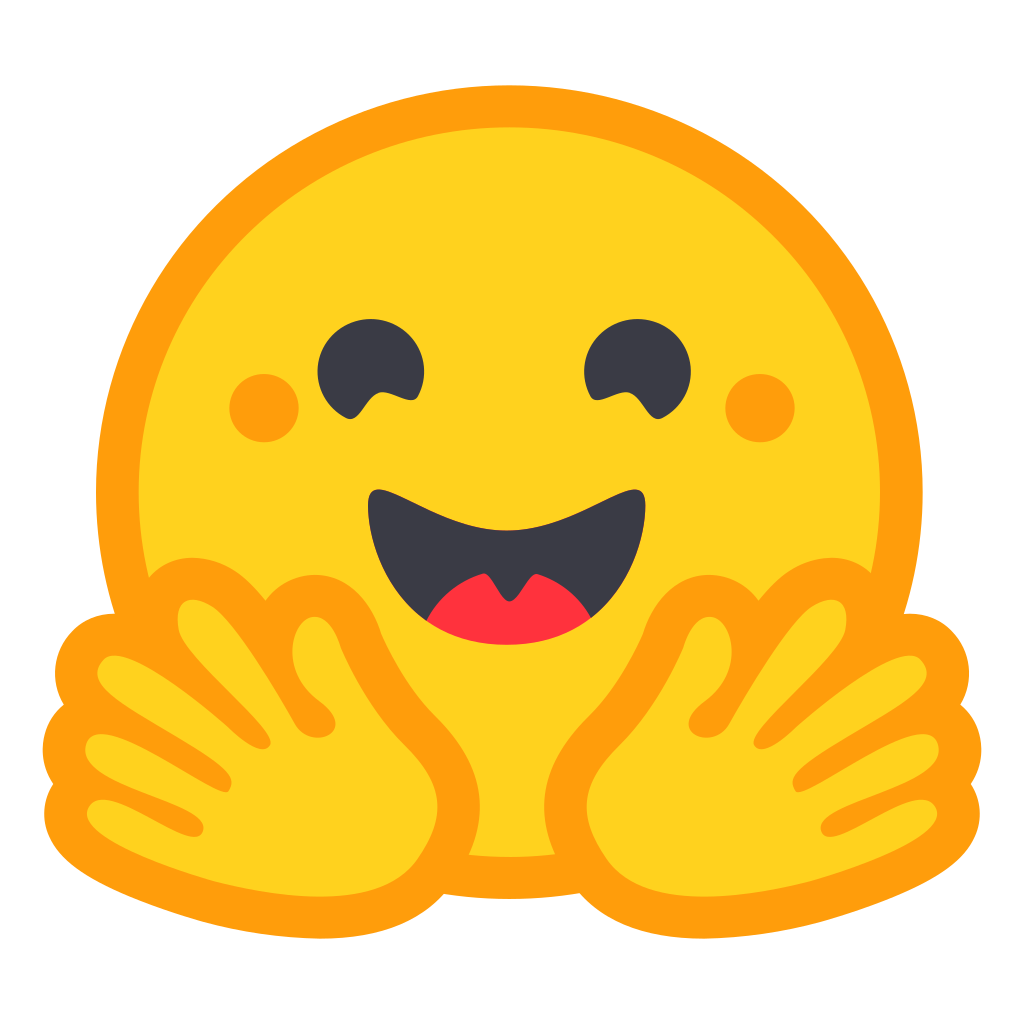}}\hspace{0.5mm}\href{https://huggingface.co/collections/cotulabs/cotu-exact-2026}{\hspace{1mm}\texttt{cotulabs/cotu-exact-2026}}
\end{center}
\vspace{-0.5em}

\end{abstract}

\section{Introduction}
\label{sec:introduction}

Large Language Models (LLMs)~\cite{brown2020language,ouyang2022instructgpt,touvron2023llama,zhao2023survey,minaee2024llmsurvey,chang2024evaluation} have advanced automated question answering, including educational and academic assistance~\cite{kamalloo-etal-2023-evaluating,wang2025llmseducation}. Despite strong natural language capabilities, purely neural models remain limited in transparency, explainability, and reliability on complex reasoning tasks~\cite{DBLP:conf/itadata/NguyenVNBNTNNBBNCLURQ25}. In academic settings, where factual accuracy is essential for student learning, concise but unsubstantiated answers or ``hallucinations''~\cite{ji2023survey,huang2025survey} from black-box models can harm pedagogy and hinder practical adoption of AI in education.

To address this challenge, \textbf{\sffamily{Neuro-Symbolic (NeSy)}}~\cite{garcez2023neurosymbolic,yang2025nesy} reasoning combines the pattern recognition and linguistic flexibility of neural networks with the deterministic guarantees of symbolic logic. This integration enables systems not only to produce accurate answers but also to expose intermediate reasoning in formal form, such as \textbf{\sffamily{First-Order Logic (FOL)}}, which can be checked with automated theorem provers like the \textbf{\sffamily{Z3 Solver}}~\cite{demoura2008z}. By making intermediate steps transparent and verifiable, neuro-symbolic systems offer a more trustworthy bridge between machine inference and human understanding.

The \textbf{\sffamily{EXACT 2026}} challenge (The 2nd International XAI Challenge for Transparent Educational Question-Answering), organized as part of the IEEE International Joint Conference on Neural Networks (\textbf{\sffamily{IJCNN 2026}}), serves as a premier venue to foster such hybrid solutions. Building on \textbf{\sffamily{EXACT 2025}}~\cite{DBLP:conf/itadata/NguyenVNBNTNNBBNCLURQ25}, this year's edition adds quantitative STEM problem-solving in physics alongside academic regulation queries, increasing demands on arithmetic precision and scientific interpretability. Furthermore, the organizers enforce strict technical constraints, including an 8-billion parameter ceiling for all LLM components and mandatory self-hosting with OpenAI-compatible tools, so participants must focus on algorithmic design and tool integration rather than proprietary models.

To guide the design and evaluation of our system, we formulate the following research questions:
\begin{itemize}
    \item[\textbf{RQ1:}] How can resource-constrained open-weight language models ($\le$ 8B parameters) perform reliable and transparent reasoning on natural language academic regulations without hallucinations?
    \item[\textbf{RQ2:}] How can symbolic logic solvers and Program-of-Thought (PoT) execution complement neural models to deliver mathematically and logically sound answers for multi-step STEM problems?
    \item[\textbf{RQ3:}] How do targeted Supervised Fine-Tuning (SFT) data curation and latency-aware serving stacks affect the inference quality and speed of a hybrid neuro-symbolic system under strict execution time budgets?
\end{itemize}

This report presents the methodology of team \cotu{}, which placed \textbf{3rd overall} in EXACT 2026 while recording the \textbf{highest final-round technical score} (\textbf{13.44/15}), combining automated answer evaluation with expert assessment of structured reasoning depth. The main contributions of this work are summarized as follows:
\begin{itemize}
    \item[\textbf{C1:}] \textbf{Unified Neuro-Symbolic PoT Pipeline:} We propose a unified, dual-task neuro-symbolic framework. For regulation queries, it translates premises to FOL and verifies them via a Z3 solver; for physics problems, it generates numerical Python code evaluated directly in a sandboxed interpreter, both bound in a shared self-correction loop.
    \item[\textbf{C2:}] \textbf{Teacher-Distillation and Adapters:} We develop a targeted data curation and SFT recipe. By distilling high-quality reasoning traces and Z3 solver code from offline teacher models, we train specialized adapters for a 4B parameter backbone to generate correct program structures and API calls.
    \item[\textbf{C3:}] \textbf{High-Throughput Serving and Empirical Success:} We deploy our system using SGLang with RadixAttention prefix caching and DFLASH speculative decoding. Our approach achieves the highest final-round technical score (13.44/15) in EXACT 2026 and a perfect score on the physics task, demonstrating the viability of hybrid reasoning on small models.
\end{itemize}

The remainder of this report is organized as follows: \Cref{sec:challenge_overview} reviews the EXACT 2026 challenge tasks and metrics. \Cref{sec:methodology} describes the proposed neuro-symbolic methodology, dataset synthesis, and fine-tuning details. \Cref{sec:experimental_setup} outlines the experimental settings and hyperparameter choices. \Cref{sec:results} presents the evaluation results and ablation studies. Finally, \Cref{sec:conclusion} concludes the paper.

\section{Challenge Overview}
\label{sec:challenge_overview}

The \textbf{\sffamily{EXACT 2026}} challenge benchmarks educational QA systems that are transparent as well as correct: every answer must be accompanied by a verifiable, step-by-step rationale to be useful in learning environments. The dataset is split into two tasks representing two core aspects of academic reasoning: logical reasoning over academic regulations and quantitative STEM problem-solving.

\subsection{Task Overview}
\label{subsec:tasks}

\paragraph{Task 1 (Logic-Based Educational Queries).}
Task 1 evaluates logical reasoning over actual university regulations and academic policies, such as grading schemes, enrollment prerequisites, scholarship criteria, and graduation requirements. Questions are formulated as Multiple-Choice (MCQ), Yes/No/Uncertain (YNU), or short Open-Ended (OE) items. At evaluation time the system receives a question together with its corresponding natural language premises; the training set additionally provides FOL translations and human-written explanations. The main difficulty lies in parsing nested rules, conditional clauses, and double negatives common in academic policies, which frequently cause standard LLMs to hallucinate. Systems must translate natural language into symbolic logic and defer to deterministic solvers for hallucination-free deduction.

\paragraph{Task 2 (Physics Problems).}
Task 2 extends the challenge into STEM domains, requiring systems to solve text-based physics problems. The problems center on electric circuits and electrostatics---equivalent resistance, voltage, current, electrical power, capacitance, electric fields, and energy---but are not limited to them: the final round also draws on broader introductory-physics topics such as mechanics, rotational dynamics, thermodynamics, and geometric optics. All items are numerical, requiring multi-step computation. Unlike Task 1, the system receives only the raw text of the question, with no premises or formula sheet. It must recover the relevant physical laws from parametric memory or external knowledge, construct the correct equations, perform sequential \textbf{\sffamily{Chain-of-Thought (CoT)}}~\cite{wei2022chain,kojima2022large} reasoning, and output the final numerical answer with its unit where applicable. Combining qualitative physical reasoning with precise arithmetic is difficult under strict model size limits, where small language models are prone to computational errors.

\subsection{Evaluation Metrics}
\label{subsec:eval_metrics}

Submissions are scored along three dimensions. \textbf{P1} measures answer correctness. For Task 2 the reference values are compared under a relative tolerance rather than by exact match: a prediction $\hat{y}$ is accepted when $|\hat{y}-y| / |y| < \tau$ against the non-zero reference answer $y$, for a threshold $\tau$ fixed by the committee, and, where the answer carries a unit, the unit must match as well. \textbf{P2} measures supporting-premise quality on Task 1, computed as the F1 between the premises a system cites and the gold supporting-premise indices; Task 2 is answer-only and carries no P2. \textbf{P3} measures the depth of structured reasoning (FOL translations, symbolic proofs, CoT traces) and is judged manually by domain experts.

The two automated rounds add a time bonus $B$ that rewards fast correct answers on top of this base score; the final round applies no such bonus. Each question has a hard time limit $T = 60$ seconds: a response that does not finish within $T$ receives $0$ points for that question. Summing over every question answered correctly ($\text{P1}_i = 1$), the bonus is
\begin{equation}
\label{eq:time_bonus}
B = r \sum_{i\,:\,\text{P1}_i = 1} \max\!\left(0,\ 1 - \frac{d_i}{T}\right),
\end{equation}
where $d_i$ is the response time for question $i$ and $r = 0.1$ is the time-bonus rate that converts each normalized time saving into score points; a quicker correct answer therefore earns a larger bonus. The per-task scores combine these components as
\begin{equation}
\label{eq:task_scores}
S_{\text{T1}} = \tfrac{1}{2}\,\text{P1} + \tfrac{1}{2}\,\text{P2} + B,
\qquad
S_{\text{T2}} = \text{P1} + B,
\end{equation}
so Task 1 weights answer correctness and premise quality equally, while Task 2 is scored on answer correctness alone.

The final round drops the time bonus and instead combines three components: the automated API score $S_{\text{API}}$ (the P1/P2 base score of \Cref{eq:task_scores} without $B$, aggregated over the round's items), the manually judged reasoning depth P3, and a presentation score $S_{\text{pres}}$. The technical score sums the first two, and the overall final score adds presentation:
\begin{equation}
\label{eq:round3_score}
S_{\text{tech}} = S_{\text{API}} + \text{P3},
\qquad
S_{\text{r3}} = S_{\text{tech}} + S_{\text{pres}} = S_{\text{API}} + \text{P3} + S_{\text{pres}},
\end{equation}
with $S_{\text{API}}$ scored out of $11$---one point per item over the $6$ Task 1 and $5$ Task 2 questions of the final round---P3 out of $4$ (so $S_{\text{tech}}$ out of $15$), and $S_{\text{pres}}$ out of $15$. The concrete evaluation rounds and per-round scores are reported in \Cref{sec:results}.

\section{Methodology}
\label{sec:methodology}







\begin{figure}[htbp]
    \centering
    \includegraphics[width=0.95\linewidth]{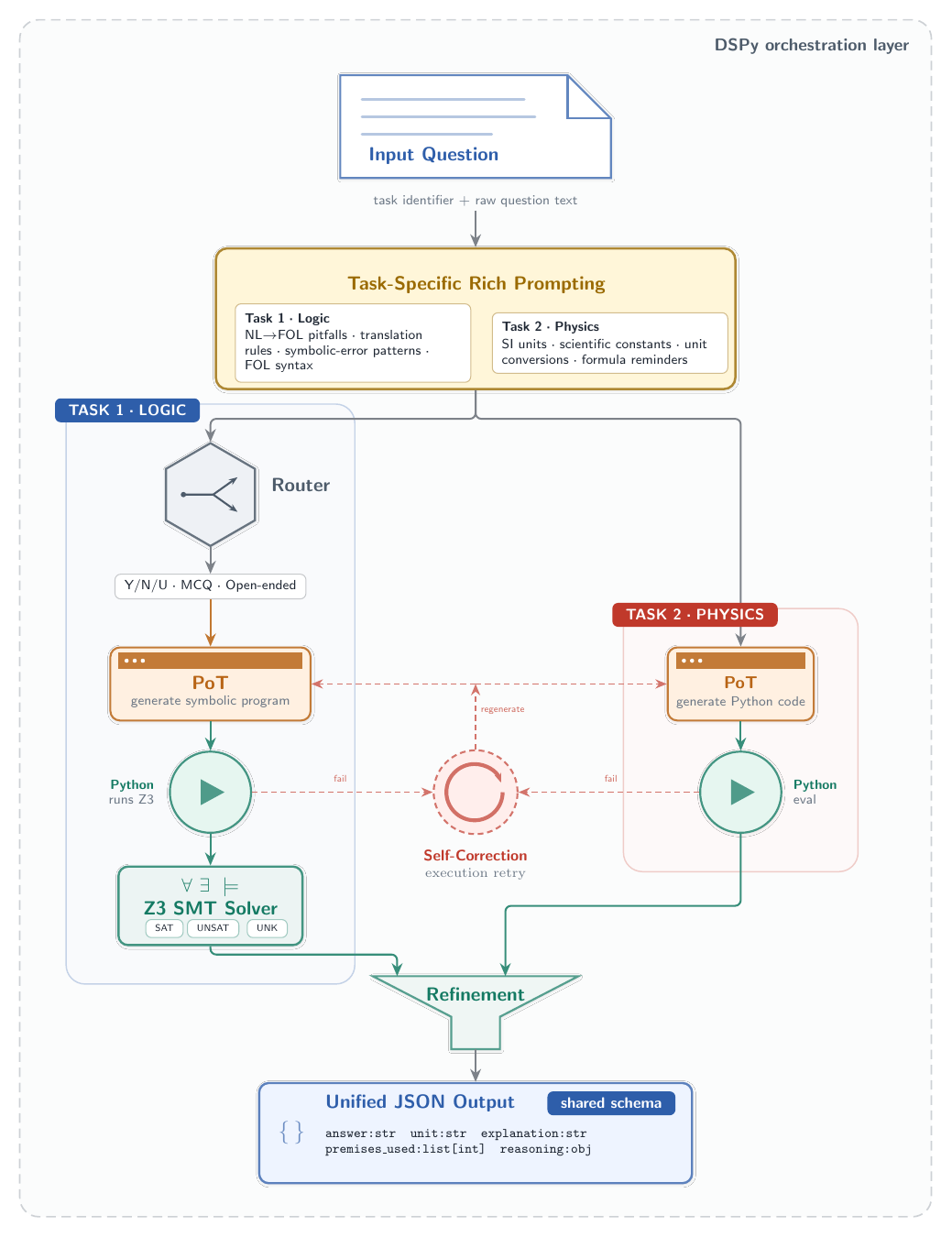}
    \caption{%
      The \cotu{} dual-task pipeline. A query, tagged with its
      task identifier, first receives task-specific rich prompting: translation
      and symbolic-error guidance for Task~1, and units, constants, and formula
      guidance for Task~2. The Task~1 branch routes the query by answer type, then
      generates a PoT that runs \textbf{\sffamily{Z3}}
      through Python, whereas the Task~2 branch generates PoT Python that is
      evaluated directly. Both branches send failed executions back through a shared
      self-correction loop that regenerates the code, and both converge on an
      output-refinement stage that emits one unified JSON record. The whole
      system is coordinated by a DSPy orchestration layer.%
    }
    \label{fig:pipeline}
\end{figure}

Our system addresses both EXACT 2026 tasks under the challenge constraints: open-source LLMs within the 8-billion parameter ceiling, self-hosted inference, and a natural language explanation accompanying every answer. We build one dedicated solver per task---Task~1 compiles regulation queries into \textbf{\sffamily{Z3}} constraints, Task~2 solves physics problems with executable Python---over a shared stack that both solvers draw on for orchestration, serving, code execution, and output formatting. \Cref{fig:pipeline} shows the full pipeline. The rest of this section covers the shared stack (\Cref{sec:method:overview}), the two solvers (\Cref{sec:method:task1,sec:method:task2}), and the data preparation and SFT behind the backbone (\Cref{sec:method:data_sft}).

\subsection{System Overview and Shared Stack}
\label{sec:method:overview}
The design principle behind \cotu{} is \emph{neuro-symbolic}: rather than
have the backbone state an answer directly, we have it write a program, and the program
produces the answer. Both branches follow this PoT pattern---Task~1 emits \textbf{\sffamily{Z3}}~\cite{demoura2008z}
constraints discharged by a symbolic solver, Task~2 emits numerical Python evaluated
directly---so the final answer is grounded in code execution rather than free-form
generation. The two branches share one orchestration, serving, and execution stack and
diverge only in what code they emit; \Cref{fig:pipeline} shows where they split and rejoin.

\paragraph{Orchestration and Backbone.} Every stage is a \textbf{\sffamily{DSPy}}~\cite{khattab2023dspy}
module, composing signatures, CoT, and
PoT steps into one declarative graph per task that we retune without rewiring the pipeline. The
backbone is \backbone{}\footnote{\url{https://huggingface.co/OrionLLM/GRM-2.5}},
an open, Qwen3.5-4B-based model fine-tuned for STEM and problem solving that fits
the 8-billion parameter ceiling; each branch loads its own SFT checkpoint (\Cref{sec:method:data_sft}). A single \textbf{\sffamily{FastAPI}} endpoint routes
each query to its branch by task \texttt{type}.

\paragraph{Rich Prompting.} A 4B model has little headroom, so we front-load each branch's prompt
with task-specific guidance that steers it toward well-formed code. Task~1 prompts carry general
logic rules and common traps---double negatives, nested conditions---alongside \textbf{\sffamily{Z3}}
API usage; Task~2 prompts carry SI-unit conventions, physical constants, unit-conversion rules,
and reminders of common physics formulas. This guidance is what lets a small model emit code that
parses and runs on the first attempt more often than not.

\paragraph{High-throughput Serving.} Serving is tuned for the challenge's 60-second per-query
limit and its time-based scoring bonus, both of which reward fast correct answers
(\Cref{subsec:eval_metrics}). We self-host with \textbf{\sffamily{SGLang}}~\cite{zheng2024sglang},
whose RadixAttention reuses the key-value cache across the shared prefixes of our prompts, and
pair it with \bsf{DFLASH} speculative decoding~\cite{chen2026dflash}, which drafts a block of tokens in
one parallel forward pass with a lightweight block-diffusion model---higher-quality drafts than
the autoregressive multi-token-prediction drafters commonly used for speculation. Together they
cut the latency of our multi-call reasoning chains, earning time bonus without shortening the
reasoning itself. We deliberately serve the backbone in BF16 with no quantization: on long-thinking,
hard problems the reasoning quality this preserves is worth more than the extra speed quantization
would buy.

\paragraph{Shared Execution and Output.} Two mechanisms close the loop for both branches. Generated
code runs in a common sandboxed interpreter---a fresh process with a restricted set of builtins and
a fixed timeout, exposing \textbf{\sffamily{Z3}} constructors on the Task~1 side and plain numerical
code on the Task~2 side---and any parse or execution failure is fed back to the backbone to
regenerate, within a fixed retry budget (\Cref{sec:experimental_setup}). Both branches then converge
on a refinement stage that populates the shared JSON schema with the fields $\langle$ \texttt{answer}, \texttt{unit}, \texttt{explanation}, \texttt{premises\_used}, \texttt{reasoning} $\rangle$, so a single
output contract serves both automated scoring and manual reasoning assessment; the
\texttt{explanation} is non-empty for every answer, as the challenge requires. This shared
self-correction loop and output-refinement stage are where the branches rejoin in \Cref{fig:pipeline}.

\subsection{Task 1: Logic-Based Educational Queries}
\label{sec:method:task1}

Task~1 answers questions about university regulations. The branch receives a question and its
natural language premises (plus answer options for multiple-choice items), encodes them as a
\textbf{\sffamily{Z3}} program, and reads the answer off the solver outcome, so the deduction
is deterministic rather than free-form. The paragraphs below detail the Task~1 branch of
\Cref{fig:pipeline}, from answer-type routing to the shared refinement stage.

\paragraph{Question Routing.} A query first passes a lightweight classifier that infers its
answer type---MCQ, YNU, or OE---from the
question text, applying MCQ only when explicit option markers are present. The type selects
one of three specialized PoT solvers, each pinned to its own answer
format. A preceding clarity check screens for malformed items, such as undefined symbols or
missing options, and short-circuits them to an \text{Uncertain} answer instead of forcing
a spurious deduction.


\paragraph{Hybrid Few-shot Retrieval.} To supply worked Z3 exemplars, we built a hybrid
retriever over a corpus of solved Task~1 items. For a query $q$ and a candidate solved item $d$ in the corpus $\mathcal{D}$, we compute dense similarity using a dense encoder (Qwen3-Embedding-0.6B) and lexical similarity using a sparse BM25 index, both served from a Qdrant store. The two ranked lists are merged using Reciprocal Rank Fusion (RRF)~\cite{10.1145/1571941.1572114}, formalized as follows:

\begin{definition}[Reciprocal Rank Fusion]
\label{def:rrf}
Let $\mathcal{R}_{\text{dense}}(q)$ and $\mathcal{R}_{\text{sparse}}(q)$ be the ranked lists of candidate items retrieved by the dense and sparse models, respectively. For each candidate item $d \in \mathcal{R}_{\text{dense}}(q) \cup \mathcal{R}_{\text{sparse}}(q)$, let $r_{\text{dense}}(q, d)$ and $r_{\text{sparse}}(q, d)$ denote the rank of $d$ in the respective lists (with $r(q, d) = \infty$ if $d$ is not present in the list). The RRF score of $d$ for query $q$ is defined as:
\begin{equation}
\label{eq:rrf_score}
\text{RRF}(q, d) = \frac{1}{k + r_{\text{dense}}(q, d)} + \frac{1}{k + r_{\text{sparse}}(q, d)},
\end{equation}
where $k$ is a constant smoothing parameter (we set $k = 60$).
\end{definition}

Top candidates with the highest $\text{RRF}(q, d)$ scores are filtered to drop near-copies of the query and injected into the PoT prompt as style and API-usage references, not logic to copy. The aim is narrow: showing the Z3 code of a similar question lets the model pattern the unfamiliar solver API and emit fewer malformed constraints. Task~2 does not use retrieval, as its Python is ordinary physics arithmetic the backbone generates reliably without an exemplar. We enabled retrieval only for the first evaluation round, so it is not part of the submitted final-round system.


\paragraph{Program-of-Thought with Z3.} The selected solver prompts the backbone to emit a
Python program that encodes the premises and query through a constrained set of Z3 helper
constructors rather than raw solver calls. The sandbox executes the program, and an entailment
routine applies a two-step unsatisfiability test using Satisfiability Modulo Theories (SMT). We formalize this verification process as follows:

\begin{definition}[SMT-based Entailment and Contradiction]
\label{def:smt_entailment}
Let $\mathcal{P} = \{p_1, p_2, \dots, p_N\}$ be a set of FoL formulas representing the natural language premises, and let $\mathcal{C}$ be the logical formula representing the query. Let $\text{Sat}(\Phi)$ denote a boolean function that returns $\text{True}$ if the logical formula $\Phi$ is satisfiable, and $\text{False}$ otherwise. The entailment status of $\mathcal{C}$ with respect to $\mathcal{P}$ is defined as:
\begin{equation}
\label{eq:entailment_status}
\text{Status}(\mathcal{P}, \mathcal{C}) = \begin{cases}
\text{Entailed}, & \text{if } \text{Sat}\left( \bigwedge_{i=1}^N p_i \land \neg\mathcal{C} \right) = \text{False}, \\
\text{Contradicted}, & \text{if } \text{Sat}\left( \bigwedge_{i=1}^N p_i \land \mathcal{C} \right) = \text{False}, \\
\text{Uncertain}, & \text{otherwise}.
\end{cases}
\end{equation}
\end{definition}

Each solver maps this status to its respective format: a Yes/No/Uncertain label for YNU, a per-option status for MCQ, and the set of entailed facts for OE. Grounding the answer in the solver verdict rather than the model's own claim is what keeps the deduction hallucination-free. We state the correctness of this symbolic grounding in the following proposition:

\begin{proposition}[Soundness of Symbolic Grounding]
\label{prop:soundness}
Assuming that the natural language premises and the query are faithfully translated into their respective FoL formulas $\mathcal{P}$ and $\mathcal{C}$, the solver verdict $\text{Status}(\mathcal{P}, \mathcal{C})$ is sound with respect to classical first-order logic semantics:
\begin{enumerate}
    \item If $\text{Status}(\mathcal{P}, \mathcal{C}) = \text{Entailed}$, then $\bigwedge_{i=1}^N p_i \models \mathcal{C}$.
    \item If $\text{Status}(\mathcal{P}, \mathcal{C}) = \text{Contradicted}$, then $\bigwedge_{i=1}^N p_i \models \neg\mathcal{C}$.
\end{enumerate}
\end{proposition}

The shared refinement stage then writes the verdict into the JSON schema and, for Task~1, emits the indices of the premises that support the answer; these are scored as the automatic premise-F1 component (P2) and therefore matter as much as the answer itself under the Task~1 metric.


\paragraph{Latency Fallback.} The main solver runs the backbone with thinking enabled at every generation step---code generation, self-correction, and output writing---which drives its accuracy but also makes a single reasoning-heavy pass liable to exceed the challenge's per-query time limit. We therefore run this solver asynchronously alongside a faster self-consistency runner that drops the thinking budget. We formalize this dual-path routing and voting fallback mechanism as follows:

Let $a_{\text{main}}$ be the answer generated by the main thinking solver, and let $d$ be its execution time. Let $t_{\text{fallback}}$ denote the safety time limit (where $t_{\text{fallback}} < T = 60$ seconds). The self-consistency runner samples $M$ independent non-thinking reasoning paths, yielding a set of candidate answers $A = \{a_1, a_2, \dots, a_M\}$, where each $a_i$ is a tuple consisting of the predicted answer value $\hat{y}_i$ and the set of cited premises $P_i$. 

The majority vote set $V_{\text{maj}}$ of the candidate answers is defined as:
\begin{equation}
  \label{eq:majority_vote}
  V_{\text{maj}} = \arg\max_{y \in \mathcal{Y}} \sum_{i=1}^M \mathbb{I}(\hat{y}_i = y),
\end{equation}
where $\mathcal{Y}$ is the space of all possible answers, and $\mathbb{I}(\cdot)$ is the indicator function. If there is a tie, i.e., $|V_{\text{maj}}| > 1$, we break the tie by selecting the candidate citing the minimum number of premises:
\begin{equation}
  \label{eq:tie_breaking}
  a^* = \arg\min_{y \in V_{\text{maj}}} \min_{i\,:\,\hat{y}_i = y} |P_i|.
\end{equation}
The final returned answer $\hat{Y}$ is chosen according to the fallback decision rule:
\begin{equation}
  \label{eq:fallback_rule}
  \hat{Y} = \begin{cases}
    a_{\text{main}}, & \text{if } d \le t_{\text{fallback}}, \\
    a^*,             & \text{otherwise}.
  \end{cases}
\end{equation}

By employing this fallback mechanism, a slow query degrades gracefully to a faster consensus answer rather than causing a timeout. The two runners thus trade accuracy for speed along a single axis---thinking versus non-thinking on the same model---and the concrete timeouts and sample count are given in \Cref{sec:experimental_setup}.

\subsection{Task 2: Physics Problems}
\label{sec:method:task2}

Task~2 solves text-based physics problems, centered on electric circuits and electrostatics but
extending to broader introductory-physics topics such as mechanics, thermodynamics, and optics.
Every item calls for a numerical answer, with a physical unit where the quantity carries one. At evaluation the branch receives only the question, with no
premises or formula sheet, so it must recover the governing laws from parametric memory and turn
them into a computation. As the Task~2 branch of \Cref{fig:pipeline} shows, it reuses the same PoT, sandbox, and self-correction machinery as Task~1, but generates plain
numerical Python that is evaluated directly instead of \textbf{\sffamily{Z3}} constraints, and uses
no retrieval, since ordinary physics arithmetic is code the backbone produces reliably without an
exemplar.


\paragraph{Program-of-Thought Solver.} The solver prompts the backbone to emit a Python program that sets up the relevant equations and computes the result through multi-step arithmetic. We formalize the Program-of-Thought (PoT) computation process for physics problems as follows:

\begin{definition}[Physics Computation]
  \label{def:pot_physics}
  Let $Q$ be a physics problem query containing a set of known input physical variables $V_{\text{in}} = \{v_1, v_2, \dots, v_m\}$ and a target quantity $y$. A PoT program $\pi$ is defined as a sequence of $L$ assignment steps:
  \begin{equation}
    \label{eq:pot_assignment}
    v_j \leftarrow \phi_j\left(\{v_i\}_{i < j},\, C_{\text{phys}}\right), \quad j = m+1, \dots, m+L,
  \end{equation}
  where $C_{\text{phys}}$ is a set of standard physical constants, $\phi_j$ is a physical formula or numerical operation, and the final assignment yields the predicted target value $\hat{y} = v_{m+L}$.
\end{definition}

The sandbox executes the program $\pi$ with numerical Python, exposing \textbf{\sffamily{SciPy}}~\cite{virtanen2020scipy} so the program can pull physical constants $C_{\text{phys}}$, such as the vacuum permittivity $\varepsilon_0$ and the elementary charge $e$, from its \texttt{scipy.constants} tables. This avoids hard-coding rounded values whose accumulated error could push the predicted value $\hat{y}$ outside the challenge's relative-tolerance band. A failure to parse or execute the sequence of assignments in \Cref{eq:pot_assignment} is fed back for regeneration under the same fixed retry budget as Task~1 (\Cref{sec:experimental_setup}). Where an answer carries a unit, the challenge scores it as part of answer correctness, so the solver emits the unit alongside the value and the shared refinement stage keeps the two together. The \texttt{premises\_used} field, which carries gold supporting-premise indices in Task~1, here records the physical laws and formulas the solution invoked; it is not scored, as the Task~2 metric is answer-only (\Cref{subsec:eval_metrics}).

\subsection{Data Preparation and Supervised Fine-Tuning}
\label{sec:method:data_sft}

The two SFT checkpoints behind the backbone (\Cref{sec:method:overview}) share a recipe: we clean
and augment the official EXACT 2026 release into a task-specific set, re-annotate it with reasoning
traces from a stronger offline teacher, and fine-tune \backbone{} to reproduce those
traces. Two teacher models are used, both
strictly offline for data preparation and never on the live solver path: DeepSeek-V4-Pro
for reasoning-trace re-annotation and Qwen3.6-27B~\footnote{https://huggingface.co/Qwen/Qwen3.6-27B} for Task~2 item synthesis. We
disclose both as non-competing, per the challenge rules.

\begin{table}[t]
  \centering
  \caption{%
    Data preparation for the two SFT sets. Task~1 is filtered by the live clarity check
    (\Cref{sec:method:task1}) and rebalanced with Z3-verified open-ended items; Task~2 is
    augmented with synthesized multiple-choice items. Synthesis teachers are offline and
    non-competing.%
  }
  \label{tab:data_prep}
  \renewcommand{\arraystretch}{1.15}
  \begin{tabularx}{\textwidth}{l X r}
    \toprule
    \textbf{Task} & \textbf{Data stage} & \textbf{Items} \\
    \midrule

    Task~1
      & \quad Official logic-query release & 808 \\
      & \quad After clarity filter (403 YNU / 314 MCQ / 5 OE) & 722 \\
      & \qquad $+$ synthesized OE (Z3-verified) & 333 \\
    \cmidrule(l){2-3}
      & \quad \textbf{Total Task 1} & 1,055 \\

    \midrule

    Task~2
      & \quad Official physics set & 1{,}352 \\
      & \qquad $+$ synthesized MCQs (Qwen3.6-27B) & 100 \\
    \cmidrule(l){2-3}
      & \quad \textbf{Total Task 2} & \textbf{1{,}452} \\

    \bottomrule
  \end{tabularx}
\end{table}

\paragraph{Task 1 Data.} We start from the official logic-query release and pass each item through
the same clarity check the live solver uses (\Cref{sec:method:task1}), keeping only the items it
labels well-formed and dropping those with missing symbol definitions or unresolved references. The
clean set is skewed by answer type toward YNU and MCQ, with only five open-ended items, so we
synthesize additional open-ended items with the teacher, discharging each candidate through
\textbf{\sffamily{Z3}} and keeping only those whose synthesized answer the solver confirms, so the
augmentation inherits the same symbolic grounding as the solver itself. \Cref{tab:data_prep}
summarizes the resulting item counts at each stage.

\paragraph{Task 2 Data.} We start from the official physics set and add multiple-choice items
synthesized from existing questions with Qwen3.6-27B, broadening
answer-format coverage beyond the numerical-only originals (\Cref{tab:data_prep}).

\paragraph{Re-annotation and Fine-tuning.} For both tasks we re-annotate every item with DeepSeek-V4-Pro, recording its thinking trace alongside the answer, explanation, and
generated program, and keep the executable trace as the fine-tuning target. We then run separate
Task~1 and Task~2 SFT of \backbone{} on these traces with LoRA~\cite{hu2022lora},
computing loss only over the assistant's reasoning and code so the model learns to produce the
structured PoT output the pipeline consumes; the trained adapters are merged back
into the backbone and released as standalone checkpoints. The goal is narrow: a 4B backbone follows the pipeline's CoT/PoT
instructions and per-task output formats far more reliably after distilling the teacher's traces
than the base model does. The two checkpoints are released as
\sftOne{}\footnote{TODO: insert link to the model} and \sftTwo{}\footnote{TODO: insert link to the model}. The exact training hyperparameters, including LoRA settings, learning rates, and epochs, are detailed in \Cref{sec:appendix_sft_hp}. Replacing the base backbone with these
checkpoints between Round 1 and Round 2 raises accuracy and cuts response time, as
\Cref{sec:results:leaderboard,sec:results:latency} report.

\section{Experimental Setup}
\label{sec:experimental_setup}

We now detail the runtime configuration under which \cotu{} produced the scores
in \Cref{sec:results}: the compliance-relevant model sizes
(\Cref{sec:setup:compliance}), the serving and decoding setup together with the
self-correction and latency-fallback budgets deferred from the methodology
(\Cref{sec:setup:serving}), and the structure of the evaluation rounds
(\Cref{sec:setup:rounds}).

\subsection{Compliance and Model Configuration}
\label{sec:setup:compliance}

All competition-eligible runs use only open-weight models within the challenge's
8-billion parameter ceiling, served on our own hardware. The ceiling applies to
the generation LLMs; non-generative LLMs, such as the embedding model used for
retrieval, do not count against it.
\Cref{tab:compliance} lists every model on the live inference path with its size,
role, and the rounds in which it ran. Every generation model derives from the 4B
backbone \backbone{}: Round 1 runs the base backbone on both tasks, while Rounds
2 and 3 run the two task-specific SFT checkpoints,
which are LoRA fine-tunes merged back into the 4B backbone and so stay within the
ceiling. The dense retriever Qwen3-Embedding-0.6B is active on the Task~1 branch
for the first round only. The two offline teachers used for data preparation,
DeepSeek-V4-Pro and Qwen3.6-27B, exceed the ceiling but never run on the live
path; we disclose them as non-competing and omit them from the table. We serve
the generation models in BF16 and enable thinking on the reasoning-heavy steps,
such as code generation\footnote{%
  Code generation is the hardest step on Task~1: the model must encode every
  natural language premise as a correct first-order-logic statement in the
  \textbf{\sffamily{Z3}} program, so a single mis-encoded premise can change the
  solver verdict.%
}, preserving the reasoning quality that long-thinking,
hard problems demand.

\begin{table}[t]
  \centering
  \caption{%
    Models on the live inference path and their compliance with the
    EXACT 2026 8B open-weight ceiling. The base backbone runs in Round 1; the two
    task-specific SFT checkpoints (\Cref{sec:method:data_sft}), LoRA fine-tunes
    merged into the same 4B backbone, run in Rounds 2--3. Offline data-preparation
    teachers are non-competing and omitted; the non-generative retriever, active
    in Round 1 only, does not count against the ceiling.%
  }
  \label{tab:compliance}
  \renewcommand{\arraystretch}{1.15}
  \begin{tabularx}{\textwidth}{l X l c c}
    \toprule
    \textbf{Role} & \textbf{Model} & \textbf{Size} & \textbf{Rounds} & \textbf{$\leq$ 8B} \\
    \midrule
    Backbone (both tasks) \quad \quad & \backboneHF{} & 4B & 1 & \checkmark \\
    Task~1 solver & \sftOne{} & 4B & 2--3 & \checkmark \\
    Task~2 solver & \sftTwo{} & 4B & 2--3 & \checkmark \\
    Retriever (Task~1) & Qwen3-Embedding-0.6B & 0.6B & 1 & \checkmark \\
    \bottomrule
  \end{tabularx}
\end{table}

\subsection{Serving and Inference}
\label{sec:setup:serving}

We serve the backbone with \textbf{\sffamily{SGLang}} and DFLASH speculative
decoding on a single NVIDIA B200 GPU for all
rounds. 
The self-correction loop shared by both branches
regenerates code on any parse or execution failure for at most two iterations
before the branch returns its best available result.
On Task~1, the latency fallback runs the
self-consistency runner in non-thinking mode with three sampled paths and supplies the answer when
the main solver---which generates with thinking enabled---does not respond within 54\,s, well
inside the $T = 60$\,s per-query hard limit.

\subsection{Evaluation Rounds}
\label{sec:setup:rounds}

The challenge ran three rounds. The two automated rounds (Round 1 and Round 2)
score P1 and P2 with the time bonus of \Cref{eq:time_bonus} and select teams for
the final round; each covers 25 Task~1 and 25 Task~2 items. Round 2 evaluates a
refined submission over the same task structure as Round 1. The final round
(Round 3) is a live evaluation among the top teams: its automated API score
$S_{\text{API}}$ combines P1 and P2 without the time bonus over 6 Task~1 and 5
Task~2 items, and it adds the manually judged reasoning-depth score P3 and a
presentation score, aggregated as \Cref{eq:round3_score}. We report Rounds 1 and
2 as automated selection rounds and Round 3 as the scored final.

\section{Results}
\label{sec:results}

We report \cotu{}'s official scores across the three evaluation rounds, then
characterize response time and the latency fallback (\Cref{sec:results:latency})
and analyze the remaining failures (\Cref{sec:results:errors}). All numbers are
official leaderboard scores under the metrics of \Cref{subsec:eval_metrics}.

\subsection{Official Leaderboard Results}
\label{sec:results:leaderboard}

\Cref{tab:auto_rounds} reports the two automated selection rounds. \cotu{} scores
a perfect Task~2 (25/25) in both Round 1 and Round 2, so its answer-only physics
solver answers all 50 items correctly under the P1 metric. Task~1 accounts for
the improvement: its combined P1+P2 score rises from 21.24/25 in Round 1 to
23.10/25 in Round 2. The principal change between the two rounds is the
model: Round 1 runs the base \backbone{} backbone, while Round 2 swaps in the
task-specific SFT checkpoints, alongside a refined
surrounding pipeline, and the score rises with them. The P1/P2 split in \Cref{tab:auto_rounds} locates the gain: premise F1
(P2) rises from 21.48 to 24.20 while answer correctness (P1) improves more
modestly from 21.00 to 22.00, so the fine-tuning helped mainly in premise
selection rather than deduction. The time bonus also improves, from 2.78 to 4.10,
as the SFT checkpoints return correct answers faster than the base backbone
(\Cref{sec:results:latency}). Both rounds
place \cotu{} well inside the qualifying band for the final round.

\begin{table}[t]
  \centering
  \caption{%
    Automated selection rounds. Task~1 is scored on answer correctness (P1) and
    premise F1 (P2), combined as $\tfrac{1}{2}\text{P1}+\tfrac{1}{2}\text{P2}$;
    Task~2 on answer correctness (P1). The time bonus score $B$ (\Cref{eq:time_bonus})
    is added on top (\Cref{eq:task_scores}). Each round covers 25 Task~1 and 25
    Task~2 items, so every column is out of 25.%
  }
  \label{tab:auto_rounds}
  \renewcommand{\arraystretch}{1.15}
  \begin{tabularx}{\textwidth}{l X c c c c c r}
    \toprule
    \textbf{Round} & & \multicolumn{3}{c}{\textbf{Task~1}} & \textbf{Task~2} \quad & \textbf{Bonus} & \textbf{Total} \\
    \cmidrule(lr){3-5}
     & & \textbf{P1} & \textbf{P2} \quad & \textbf{P1+P2} & \textbf{P1} & \textbf{$B$} \quad & \textbf{(Rank)} \\
    \midrule
    Round 1 & Selection round 1 & 21.00 & 21.48 \quad & 21.24 & 25.00 & 2.78 \quad & 49.02\ (4/50) \\
    Round 2 & Selection round 2 & 22.00 & 24.20 \quad & 23.10 & 25.00 & 4.10 \quad & 52.20\ (6/35) \\
    \bottomrule
  \end{tabularx}
\end{table}

The final round (Round 3) drops the time bonus and adds manual reasoning and
presentation scoring. \Cref{tab:final_round} breaks down
\cotu{}'s result. The automated API score is 10.24/11, which with a reasoning-depth score of 3.20/4 gives a technical score
$S_{\text{tech}} = 13.44/15$---the \textbf{highest} of any team, ahead of the runner-up's
12.51/15. The presentation score of 8.80/15 brings the overall final score to
22.24, placing \cotu{} \textbf{3rd overall}. The technical lead and the 3rd-place
finish are therefore separate facts: \cotu{} produced the strongest system on the
automated and expert-judged reasoning metrics, while the presentation component,
which weighs equally with the full technical score, moved the overall ranking.

\begin{table}[t]
  \centering
  \caption{%
    Final round (Round 3) score breakdown, over 6 Task~1 and 5 Task~2 items. The
    technical score $S_{\text{tech}}$ sums the automated API score and the manual
    reasoning score P3; the overall score adds presentation
    (\Cref{eq:round3_score}). \cotu{}'s $S_{\text{tech}}$ is the highest among all
    teams.%
  }
  \label{tab:final_round}
  \renewcommand{\arraystretch}{1.15}
  \begin{tabularx}{\textwidth}{X l r}
    \toprule
    \textbf{Component} & & \textbf{\cotu{}} \\
    \midrule
    Automated API & $S_{\text{API}}$ (max. 11 pts) & 10.24 \\
    Reasoning depth & P3 (max. 4 pts) & 3.20 \\
    \cmidrule(l){1-3}
    \textbf{Technical} & $S_{\text{tech}}$ (max. 15 pts) & \textbf{13.44} \\
    \quad {\footnotesize runner-up technical} & {\footnotesize (max. 15 pts)} & {\footnotesize 12.51} \\
    Presentation & $S_{\text{pres}}$ (max. 15 pts) & 8.80 \\
    \midrule
    \textbf{Overall} & $S_{\text{r3}}$ & \textbf{22.24\ (3rd)} \\
    \bottomrule
  \end{tabularx}
\end{table}

\subsection{Response Time and Latency Fallback}
\label{sec:results:latency}

The EXACT time bonus rewards answering quickly, but only on items answered
correctly, and the serving stack (\Cref{sec:setup:serving}) was built to stay
well inside the $T = 60$\,s per-query limit. \Cref{tab:latency} reports mean per-query response time from the
official run logs, alongside how often the Task~1 latency fallback emitted the
self-consistency runner's result in place of the main solver.

Three patterns stand out. First, mean response time drops sharply from Round 1 to
Round 2 on both tasks: Task~1 from 32.51\,s to 12.05\,s and Task~2 from 17.20\,s
to 4.21\,s. Round 1 ran the base \backbone{} backbone while Round 2 ran the
task-specific SFT checkpoints, so the drop is
consistent with the SFT checkpoints being faster. A plausible mechanism is that
fine-tuning on reasoning traces distilled from the stronger DeepSeek-V4-Pro
teacher steers the model onto shorter, more direct
solution trajectories, so it settles on an answer with less intermediate
reasoning. The two rounds use different item sets and Round 2 also refined the
surrounding pipeline, so we read this as supporting evidence rather than a
controlled measurement, which would require holding pipeline and items fixed
while swapping only the model. Second, response time
rises again in the final round (18.00\,s on Task~1 and 10.63\,s on Task~2) even
though Round 3 runs the same SFT checkpoints as Round 2. The final round is a
smaller, harder live evaluation, and the accuracy drop reported in
\Cref{sec:results:errors} is consistent with this, so the longer times point to
items that require more reasoning effort to solve rather than a regression in the
solver. Third, the latency fallback was needed only in
Round 1, where it fired on 5 of 25 Task~1 items (20\%). In Rounds 2 and 3 no query
reached the 54\,s fallback threshold---the slowest Task~1 item in Round 2 finished
in 18.4\,s---so the fallback went from a regularly exercised safety net to an
inactive one. It is a Task~1-only mechanism by design and never triggered on Task~2. The
faster responses also lifted the time bonus, from 2.78 to 4.10,
since it rewards correct answers returned quickly.

\begin{table}[t]
  \centering
  \caption{%
    Per-query response time and latency-fallback usage across the three rounds,
    from the official run logs. Mean time is over the 25 Task~1 and 25 Task~2
    items in each automated round and the 6 Task~1 and 5 Task~2 items in the final
    round. ``Fallback'' counts Task~1 queries whose result came from the
    self-consistency fast runner after the main solver exceeded the 54\,s budget;
    it is Task~1-only by design and never fired on
    Task~2. The time bonus score $B$ is not awarded in the final round.%
  }
  \label{tab:latency}
  \renewcommand{\arraystretch}{1.15}
  \begin{tabularx}{\textwidth}{l X c c c r}
    \toprule
    \textbf{Round} & & \multicolumn{2}{c}{\textbf{Mean time (s)}} & \textbf{Fallback} & \textbf{Bonus} \\
    \cmidrule(lr){3-4}
     & & \textbf{Task~1} \quad & \textbf{Task~2} \quad & \textbf{Task~1} & \textbf{$B$} \\
    \midrule
    Round 1 \quad & Base backbone & 32.51 & 17.20 & 5/25 & 2.78 \\
    Round 2 \quad & SFT checkpoints & 12.05 & 4.21 & 0/25 & 4.10 \\
    Round 3 \quad & SFT checkpoints & 18.00 & 10.63 & 0/6 & --- \\
    \bottomrule
  \end{tabularx}
\end{table}

\subsection{Error Analysis}
\label{sec:results:errors}

We inspect the per-query logs from all three rounds to characterize where \cotu{}
still fails. Across every round, all incorrect items fall on Task~1: the Task~2
solver is correct on every automated- and final-round item. The failures are
therefore confined to Task~1, and they split into two modes matching the Task~1
metric: \emph{wrong answer}, where the deduction
itself is incorrect, and \emph{wrong premises}, where the answer is correct but the
cited supporting premises miss the gold set and lose P2.

In Round 1, 36 of 50 items are fully correct; of the 14 Task~1 misses, 10 are
premise-selection errors and 4 are wrong answers, so most of the lost Task~1
score is premise F1 (P2) rather than deduction. Round 2 shows the effect of the
refined submission: 45 of 50 correct, with only 2 premise errors and 3 wrong
answers on Task~1. In the final round, the 6 Task~1 items break down into 2
correct, 3 premise-selection errors, and 1 wrong answer, so premise selection
remains the dominant Task~1 failure mode into the final round.

That the leading error mode is premise selection rather than deduction is
consistent with the design: grounding the answer in the Z3 solver verdict
(\Cref{sec:method:task1}) keeps deductions largely correct, while identifying the
exact supporting premises---scored as P2 and weighted equally with answer
correctness under $S_{\text{T1}}$---remains the harder subproblem.

\section{Conclusion}
\label{sec:conclusion}

In this work, we presented team \cotu{}'s submission to EXACT 2026, a neuro-symbolic system for
transparent educational QA under the challenge's open-weight, 8B-parameter ceiling. Instead of
answering directly, the \backbone{} backbone writes a program
that is executed to obtain the answer: a Z3 encoding whose entailment verdict grounds each
regulation query in Task 1, and numerical Python for the physics problems in Task 2, with both
branches sharing a self-correction loop and a unified explained-JSON output. The system achieved
a perfect Task 2 score in both automated rounds and obtained the highest final-round technical
score of any team ($S_{\text{tech}}=13.44/15$); with the equally weighted presentation score
included, \cotu{} placed 3rd overall. Grounding deduction in a symbolic solver kept answers correct
and verifiable at the 4B scale, leaving premise selection, rather than the deductions themselves, as
the main remaining source of error.

\paragraph{Limitations.}
The main residual error mode is premise selection (P2) rather than answer correctness: the Z3
verdict keeps Task 1 deductions largely correct, but identifying the exact supporting premises,
weighted equally with the answer, remains the harder subproblem. Our data pipeline also depends on
two offline teacher models---DeepSeek-V4-Pro for reasoning-trace re-annotation and Qwen3.6-27B for
Task 2 item synthesis---that exceed the 8B ceiling. Both are disclosed as non-competing and never
enter the live inference path, yet the quality of the fine-tuned checkpoints still rests on this
distillation.

\paragraph{Future work.}
The most immediate priority is premise selection. Tighter attribution between the Z3 encoding and
the cited premise indices, or joint optimization of deduction and citation, would target the error
mode that accounts for most of our lost P2. A second direction is to reduce the reliance on large
offline teachers by generating reasoning traces on-policy from the backbone itself and retaining
only the Z3 and Python programs whose executions verify. Our pipeline already runs the solver that
provides this correctness signal, the same signal that three lines of recent work build a training
loop around: rejection-sampling and self-taught fine-tuning, which keep model-generated solutions
that pass a check~\cite{zelikman2022star,singh2024restem,yuan2023rft};
execution-feedback self-training for programs, which filters candidates by running them against
synthesized tests~\cite{wei2024selfcodealign}; and reinforcement learning with
verifiable rewards, which optimizes the policy directly against a deterministic correctness signal
rather than distilling filtered traces~\cite{deepseekai2025r1,zeng2025acecoder,yu2025dapo}. Finally, extending symbolic grounding to Task 2, through dimensional and unit
checking or verification of retrieved formulas, would carry the deterministic guarantees that
anchor Task 1 to the physics branch as its domain widens.

\bibliographystyle{splncs04}
\bibliography{refs}

\clearpage
\appendix

\section{Supervised Fine-Tuning Hyperparameters}
\label{sec:appendix_sft_hp}

\begin{table}[t]
  \centering
  \caption{LoRA configuration used to produce the Task~1 and Task~2 \cotu{}
  checkpoints described in \Cref{sec:method:data_sft}.}
  \label{tab:sft_hp}
  \begin{tabular}{lll}
    \toprule
    Hyperparameter & Task~1 (\text{T1-SFT}) & Task~2 (\text{T2-SFT}) \\
    \midrule
    Base model            & \backboneHF{} & \backboneHF{} \\
    Method                & LoRA & LoRA \\
    LoRA rank $r$         & 16 & 16 \\
    LoRA $\alpha$         & 32 & 32 \\
    LoRA dropout          & 0.0 & 0.0 \\
    Rank-stabilized LoRA  & \texttt{True} & \texttt{True} \\
    LoRA target modules   & \multicolumn{2}{l}{\texttt{q\_proj, k\_proj, v\_proj, o\_proj,}} \\
                          & \multicolumn{2}{l}{\texttt{gate\_proj, up\_proj, down\_proj}} \\
    Learning rate         & $2\times10^{-4}$ & $1\times10^{-4}$ \\
    LR schedule           & cosine & cosine \\
    Warmup                & 0.03 & 0.03 \\
    Batch size / device   & 1 & 2 \\
    Grad.\ accumulation   & 4 & 4 \\
    Epochs                & 1 & 1 \\
    Weight decay          & 0.01 & 0.01 \\
    Optimizer             & \texttt{adamw\_torch\_fused} & \texttt{adamw\_torch\_fused} \\
    \bottomrule
  \end{tabular}
\end{table}

We fine-tune a separate \backboneHF{} checkpoint for each task with
LoRA~\cite{hu2022lora}. Adapters are applied to
all attention and MLP projection matrices (\Cref{tab:sft_hp}) and use the
rank-stabilized parameterization, which rescales the update by
$\alpha/\sqrt{r}$~\cite{kalajdzievski2023rslora} to keep the effective learning
rate stable. The loss is computed
only over the model's reasoning trace ($\langle\texttt{think}\rangle\ldots
\langle/\texttt{think}\rangle$) and its final response, with the prompt tokens
masked out. After training, the LoRA adapters are merged into the base weights
and released as standalone checkpoints. The two
tasks share the configuration in \Cref{tab:sft_hp} and differ only in learning
rate and per-device batch size.

\end{document}